# PB-IAD: Utilizing multimodal foundation models for semantic industrial anomaly detection in dynamic manufacturing environments

**Bernd Hofmann[a,*], Albert Scheck[a], Jörg Franke[a], Patrick Bründl[a]**

[a] Institute for Factory Automation and Production Systems (FAPS)

Friedrich-Alexander-Universität Erlangen-Nürnberg, Germany

*Corresponding author

*E-mail address: bernd.hofmann@faps.fau.de

*ORCID: https://orcid.org/0009-0009-0666-6149

## Abstract

The detection of anomalies in manufacturing processes is crucial to ensure product quality and identify process deviations. Statistical and data-driven approaches remain the standard in industrial anomaly detection, yet their adaptability and usability are constrained by the dependence on extensive annotated datasets and limited flexibility under dynamic production conditions. Recent advances in the perception capabilities of foundation models provide promising opportunities for their adaptation to this downstream task. This paper presents PB-IAD (Prompt-based Industrial Anomaly Detection), a novel framework that leverages the multimodal and reasoning capabilities of foundation models for industrial anomaly detection. Specifically, PB-IAD addresses three key requirements of dynamic production environments: data sparsity, agile adaptability, and domain user centricity. In addition to the anomaly detection, the framework includes a prompt template that is specifically designed for iteratively implementing domain-specific process knowledge, as well as a pre-processing module that translates domain user inputs into effective system prompts. This user-centric design allows domain experts to customise the system flexibly without requiring data science expertise. The proposed framework is evaluated by utilizing GPT-4.1 across three distinct manufacturing scenarios, two data modalities, and an ablation study to systematically assess the contribution of semantic instructions. Furthermore, PB-IAD is benchmarked to state-of-the-art methods for anomaly detection such as PatchCore. The results demonstrate superior performance, particularly in data-sparse scenarios and low-shot settings, achieved solely through semantic instructions.

## Keywords

LLM, VLM, Industrial AI, Quality Control, Quality Management, Human-machine collaboration

## Introduction

Ensuring and maintaining product quality is a central element in manufacturing industries [1]. The primary objective is to fulfil customer expectations concerning product quality characteristics, as deviations lead to internal and external failure costs and complaints [2]. However, industrial processes are inherently subject to variability due to deterioration caused by wear and tear. Furthermore, process variability can be attributed to differences in raw materials, fluctuations in the performance and operation of manufacturing equipment, and inconsistencies in the execution of tasks by operators [2,3]. The early detection of such deviations is crucial in order to prevent further escalation and financial losses. In this context, anomaly detection (AD) techniques have been widespread in order to identify anomalous conditions and support short term intervention [3]. Since the variability of physical, sensory, or time-related quality characteristics can only be described in statistical terms, statistical methods play a central role in this regard [2]. In addition, data-driven approaches such as deep learning (DL) have demonstrated strong performance in applications including predictive maintenance and AD, leading to a growing shift toward their adoption [4,5]. Nevertheless, accelerated product life cycles and an increased time-to-market pressure have intensified the dynamism and complexity of industrial environments, resulting in highly uncertain and continuous ramp-up conditions [6]. These dynamic conditions also impact the initialization of industrial anomaly detection (IAD) systems. Tool wear or the integration of new product variants on the production line can lead to changes in the underlying data distribution and statistical properties relevant to the detection logic. This phenomenon, known as concept drift, results in a degradation in model performance [7]. Traditional AD models in industrial settings often lack the flexibility and adaptability required to cope with such variability, particularly when new defect types or operational changes emerge [8].

Recent advancements in generative artificial intelligence and large language models (LLM) have opened up promising opportunities across a range of domains, including IAD [9–12]. These models, which have been pretrained on extensive datasets and broad knowledge corpora, demonstrate the ability to handle complex language and vision-based tasks, with the potential to influence numerous professions and industries through their multimodal capabilities [9,13,14]. However, despite their strong reasoning abilities, they often lack the specific domain knowledge required for downstream tasks such as AD [15]. Furthermore, common adaptation strategies such as fine-tuning depend on large volumes of data, which is both costly and, particularly in dynamic industrial environments, difficult to obtain due to the inherently rare nature of appearance [13].

In order to address the challenges associated with the initialization of IAD systems in dynamic environments, the Prompt-based Industrial Anomaly Detection (PB-IAD) framework is proposed. As illustrated in Fig. 1, this approach is centered on semantic interaction and relies entirely on in-context learning (ICL) to handle data sparsity. As a result, the framework enables rapid deployment and adaptation solely through prompt engineering. Furthermore, domain experts without prior data science expertise are capable of independently configuring and adjusting the system through the utilization of natural language instructions. An ablation study and a benchmarking analysis against established machine learning (ML) algorithms for AD are conducted within three industrial scenarios to evaluate the proposed instruction template and the overall framework.

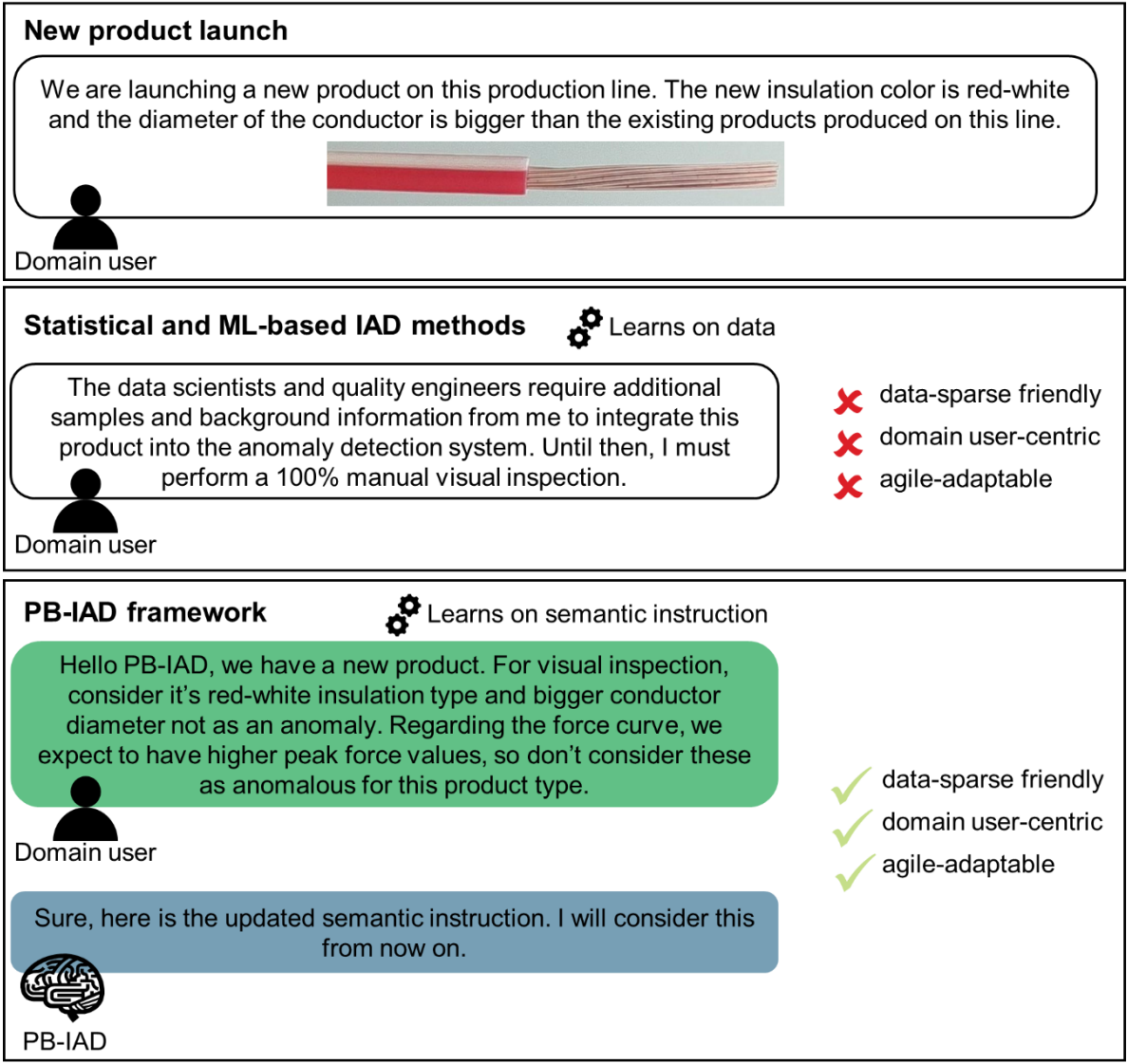

**Fig. 1.** Comparison of PB-IAD with conventional industrial anomaly detection methods in handling key challenges in dynamic production environments, such as data sparsity, domain user centricity, and adaptability to change.

# Background

## Industrial anomaly detection

In industrial context, AD is employed to identify discrepancies in products or processes at an early stage, thereby helping to prevent escalating failure costs as the process progresses [16]. An anomaly is generally defined as a significant deviation from expected behaviour or typical patterns, and it can occur across various types of data [12]. In the scope of quality control, anomalies are associated with the manufactured product, whereby they can be detected through different methodologies. In product quality control, deviations are directly observed on the product, such as irregularities in visual appearance or unexpected values in geometric or mechanical measurements. In process control, product-related anomalies are identified indirectly by observing the production process through sensor data or machine parameters, often recorded as numerical values or time series. These process signals can indicate underlying product issues, even in the absent of a visible defect. Furthermore, the analysis of metadata, including production schedules, material types, supplier information, machine conditions, and event logs can uncover atypical patterns that may significantly impact quality. In addition to the function of ensuring product quality, AD is also employed to support predictive maintenance tasks, for example the identification of unusual behaviour that may be indicative of future equipment failure. Overall, IAD is intended to ensure the consistent quality of the product and to facilitate stable production.

The detection of such outliers is defined as an ill-posed problem, with the solution, such as a threshold, depending heavily on the quality and consistency of the input data. Minor variations in the input data can have a substantial impact on the decision rule. Moreover, within the domain of manufacturing, the delineation of an anomaly frequently poses challenges due to its lack of universal clarity and the absence of a clearly defined boundary. It is challenging to determine with certainty what qualifies as an anomaly and what may still be considered an acceptable variation from normal behaviour. For instance, in surface inspection tasks, it can be of subjective nature whether a scratch or streak is sufficiently pronounced to be classified as a defect. Consequently, the solution to the AD task is inherently problem-specific and context-dependent.

## Methodologies for anomaly detection tasks in quality control

Statistical models and ML approaches have been the dominant methods for detecting anomalies in industrial processes and products.

Statistical methods such as statistical process control (SPC) charts or interquartile range (IQR) are frequently employed to monitor process parameters in production lines [17]. Deviations from the non-anomalous state are defined by upper and lower control limits, which are determined by simple statistical considerations, such as $\pm 3\sigma$, with $\sigma$ being the known or estimated standard deviation of the population [2,18]. When a process parameter exceeds the established control limits, the process is considered to be out of control, and corrective actions must be initiated.

ML methods are increasingly utilized in industrial context due to their proven efficacy in tasks like predictive maintenance and AD, by automatically learning and extracting relevant features from complex data [4,10]. The choice of learning approach depends on the type and availability of data and can be categorized as unsupervised or supervised learning. In unsupervised learning tasks, no labelled examples are available. The model is primarily trained on normal instances, learning the patterns that characterize standard behaviour, and subsequently flags deviations from these patterns as potential anomalies. If normal and anomalous samples are available and labelled for model training, the task is defined as a supervised learning problem and is typically treated as a binary classification scenario. [19]

Although traditional ML approaches have demonstrated robust performance in numerous industrial applications, recent advancements in foundation models (FM), particularly LLMs with multimodal capabilities, are opening up new possibilities for intelligent systems. These models have achieved a high proficiency in a variety of natural language processing (NLP) and perception tasks, resulting in a significant increase in research activity and practical applications across various domains [13]. State-of-the-art (SOTA) FM, such as GPT-4 [20], have been shown to be able to process various types of sequential data, are capable of generating text, images, audio, or programming code, can be equipped with external tools and communicate with other specialised models in agentic frameworks. When incorporated into software systems or utilized as standalone tools, these models can enhance, accelerate, or improve the quality of outputs in existing workflows [9]. As demonstrated in the study of Eloundou et al. [9], this phenomena is particularly evident in occupations such as mathematicians, tax preparers, financial analysts, data managers, and news analysts. These occupations have been shown to exhibit high exposure to FM's capabilities. It is evident that these roles have strong reliance on skills that are also fundamental to data analysis, including numerical reasoning, statistical evaluation, and interpretation of structured data. Consequently, FMs are increasingly applied to tasks such as time series analysis [11,12] and computer vision [14]. As a result, FMs are also gaining relevance in the domain of IAD, a field in which data analysis and visual inspection play a central role. Moreover, the capabilities of SOTA FMs enable IAD systems to move beyond the identification of outliers. Recent work of [21] proposes the utilization of FMs as industrial quality inspectors for visual inspection, expanding the scope of traditional AD to tasks such as defect localization, description and analysis.

## Prerequisites for the initialization of an IAD system in dynamic environments

A globalized economy and demand-driven markets compel companies to introduce new products with increasing frequency. As a result, production systems and supply chains are constantly being adapted, resulting in significant operational challenges and recurring ramp-up conditions [22]. Beside the influence of inherent process variability, product quality is additionally affected by several factors, including a generally lower skill and process knowledge level among operators and engineers, fatigue from overtime, schedule pressure, and inadequate working conditions [6]. Consequently, the organizational maturity of the production system and the effectiveness of the tools and equipment in use is not optimal [6]. Additionally, there is a strong dependency on individuals whose expertise is largely based on implicit knowledge that is difficult to share or formalize [23]. As a result, modern manufacturing systems are of dynamic nature and must be able to handle frequent changes in materials, products, tools, and skill levels. To remain effective under these conditions, quality control systems need to be agile-adaptable, easy to implement, and data-sparse friendly.

Quality control systems utilize different approaches, each requiring specific prerequisites to be effectively implemented. The initialization of statistical methods generally necessitates a dataset that ensures statistical significance and adequately represents the underlying process. For instance, during the initialization phase of SPC charts, small batches of newly introduced products are evaluated [17]. Typically, 25 subgroups of size 4 or 5 are considered sufficient to provide preliminary estimates [18]. The identification of outliers, characterised by points that exceed the established lower and upper control limits, initiates a systematic investigation into the underlying causes. Afterwards, the process and control lines are adjusted adequately to mitigate the impact of the identified issue. This iterative cycle of inspection and chart refinement is continued until the process reaches a stable state [17].

ML algorithms similarly necessitate a dataset for initialization, as they derive patterns directly from empirical data. These models are generally not effectively designed for incremental learning and frequently necessitate comprehensive retraining when confronted with novel data patterns. In a manufacturing context, a novel data pattern can already result from minor influences on the production equipment, such as a shift in lighting conditions or dirt on a camera lens. Additionally, ML models have limited capacity for transferring prior process knowledge, and their performance is hindered by the inconsistent availability and structure of data in new scenarios [24]. Furthermore, class imbalance remains a persistent challenge, especially in manufacturing where the proportion of defective to non-defective parts can be as low as one in a million [24].

In the domain of DL, transfer learning approaches have been demonstrated to be effective in addressing the challenges posed by limited data. This is achieved through the transfer of knowledge from a related source domain, where data is more abundant, to a target domain with limited data availability [25]. Nevertheless, the presence of diverse domain data that possesses statistical significance and adequately represents the underlying process remains a necessary prerequisite.

Beside data collection, an effective initialization of a quality control system additionally necessitates the incorporation of relevant process experience. This encompasses the selection of causal features or the definition of heuristic thresholds, based on prior knowledge of experienced domain experts. A significant challenge comes from the inherent difficulty of formalising such tacit expertise [26]. Consequently, the initialization of quality control systems measures based on statistical models or ML algorithms remains challenging in dynamic and rapidly changing production environments.

In contrast, FMs offer strong generalisation capabilities and support multi-task learning, enabling them to adapt more flexibly to varying conditions and tasks. Large-scale pretraining enables FMs to support a wide range of general-purpose tasks [13,27]. Nevertheless, a significant limitation lies in their restricted performance on tasks necessitating extensive domain-specific knowledge or substantial amounts of task-specific data [13]. Consequently, these models frequently function as a foundation for more specialised applications and can be adapted through various strategies to better suit specific use cases [28]. As a result, the general knowledge embedded in FMs can be utilized and transferred to related tasks with only minor retraining.

A common approach is fine-tuning, which involves adjusting a selected set of model parameters using task-specific datasets that were not part of the pretraining [29]. In addition to full fine-tuning, where all model parameters are updated, other strategies have been developed, such as adapter-based methods [30] and low-rank adaptation (LoRA) [31]. However, fine-tuning still requires significant datasets and computational resources [32,33].

Another methodology that can be employed is ICL, in which the model performs novel tasks by incorporating data examples or relevant semantic instructions directly within the prompt [33]. This strategy does not necessitate the modification of the model's parameters. Instead it is based on prompting techniques to guide the model's behaviour [28]. A prompt is defined as the input to the model, which it subsequently generates an output for [34]. FMs with multimodal capabilities can be provided with semantic instructions and data examples, like images. Depending on the number of examples provided at inference time the prompting strategy can be classified as zero-shot, one-shot, or few-shot [33].

In context of IAD, a semantic instruction enables the definition of quality states solely through natural language. Therefore, it facilitates a more effective approach to the inherently ill-posed problem of AD, as the quality control system can be explicitly instructed with semantic boundary rules. For instance, the model is instructed that a minor, purely cosmetic streak on the product surface is still acceptable. This reduces the necessity to collect a large dataset of such an anomaly, which may not yet be available. Additionally, ICL represents the sole strategy for adapting FM to domain-specific tasks that transfers the responsibility of providing domain knowledge from the data scientist to the process expert [28]. Therefore, the approach is particularly well-suited for user-centric systems.

As a summary, in dynamic production environments such as the ramp-up phase, where data is scarce, conditions change rapidly, and process knowledge is often tacit, ICL offers a uniquely flexible and straightforward solution for initializing or updating AD systems. Fig. 2 summarizes this chapter by schematically illustrating distinct learning approaches of AD techniques under ramp-up conditions. The figure emphasizes the initialization requirements of the techniques, particularly their reliance on accumulated data and process knowledge.

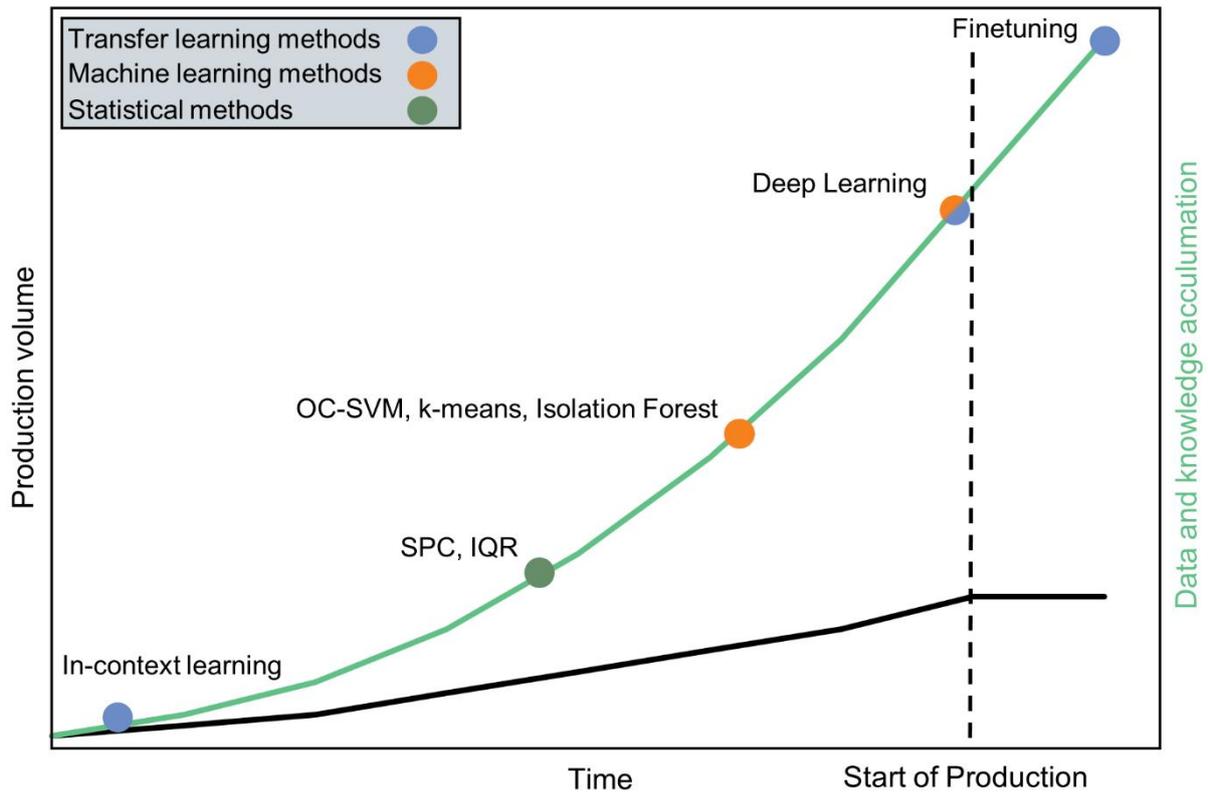

**Fig. 2.** Schematic illustration of possible deployment times of anomaly detection methods in a ramp-up scenario depending on the level of required data and process knowledge accumulation.

## Related work

Recent progress in the development of LLMs has created new opportunities for IAD. FMs are being adapted to perform domain-specific tasks like outlier detection or root cause analysis. Due to their extensive pretraining and advanced reasoning capabilities, these models offer substantial advantages in quality control, particularly in industrial environments where anomalous data is limited and manufacturing conditions change frequently. The following section reviews current frameworks that leverage FMs for IAD, with a particular focus on how they address the challenges posed by data scarcity and the dynamic nature of modern production environments.

One research direction is to perform additional training on models to enrich their capabilities on domain-specific tasks. AnomalyGPT [15], for example, fine-tunes a vision language model (VLM) on non-anomalous training and simulated defect images paired with textual descriptions to detect and localize anomalies. The framework enables multi-turn dialogues, answers questions about the anomaly and provides pixel-level localisation masks to the user.

In the time series domain, Qaid et al. [35] propose FD-LLM, a framework that encodes vibration signals as transformed frequency-domain representations and as statistical summaries. Open-source LLMs are fine-tuned on these representations and utilize instruction prompts that provide machine and operational context to classify machine fault types. Their results demonstrate strong generalization across varying operating conditions, in many cases outperforming traditional ML and DL models.

Moreover, FaultGPT [36] by Chen et al. is a VLM that generates detailed diagnostic reports directly from time-frequency images of bearing vibration signals. The Multi-Scale Cross-Modal Image Decoder, Adapter, and Prompt Learner modules are instruction-tuned on a large fault diagnosis Q&A dataset, enabling the system to produce human-readable fault analyses. The results show a significant drop in performance when instruction tuning is omitted, underscoring the importance of task-specific tuning for generalization in industrial fault diagnosis tasks.

The Myriad model [37] follows a different approach by integrating outputs from pre-trained vision-based anomaly detectors ("vision experts") into a FM. During training, anomaly maps generated by these vision experts are used to guide the attention of Myriad's vision encoder and prompt generation modules toward defective regions, thereby enhancing the model's ability to detect and describe anomalies.

Similarly, Yang et al. [38] leverage a FM as a post-processor to inspect the decision of a pretrained baseline anomaly detector to identify false alarms. By monitoring time-series plots and contextual descriptions in a zero-shot setting the model reduces the necessity of human oversight of anomaly detection systems and is able to reduce the false alarm rate significantly.

Chen et al. propose Echo [8], an agentic framework that augments a FM with several expert modules to provide the necessary industrial context. Echo leverages resources from the MMAD benchmark [21], which includes comprehensive descriptions of defects for anomaly datasets. In their approach, a reference extractor retrieves similar normal examples as a comparable baseline, a knowledge guide injects relevant defect descriptions and standards, a reasoning expert applies step-by-step logic to interpret complex cases and a final decision module synthesizes all inputs. This agentic approach employs retrieval augmented generation (RAG) and context-specific knowledge extraction without modifying the model parameters of the FM. While Echo does not require parameter adjustment of the backbone models, it does depend on a well-curated knowledge base such as the MMAD descriptions to enable precise and context-aware reasoning.

An alternative research stream concentrates on the issue of limited data availability in environments where anomalies are rare and difficult to collect. One approach is the generation of synthetic training data.

Singh et al. [39] propose to treat FM as generative time-series models by fine-tuning them on manufacturing process instructions, enabling it to output realistic sensor data sequences. Their approach produced synthetic signals that not only improved the performance of baseline AD algorithms, but also consistently outperformed traditional data generation techniques such as ARIMA and LSTM in statistical similarity.

Another promising pathway involves leveraging the pretrained knowledge by adapting FMs solely through ICL techniques, such as low-shot learning. Instead of relying on task-specific training on existing domain-specific datasets, this approach enables models to perform AD tasks by conditioning on only few or even no examples provided at inference time.

WinCLIP [40] extends the CLIP [41] VLM to handle anomaly classification and segmentation in a zero- or few-shot setting, without requiring task-specific retraining. By combining compositional language prompts with multi-scale image features, WinCLIP is able to identify visual defects using minimal or even no normal reference images. This approach consistently outperforms prior methods on standard benchmarks and demonstrates strong flexibility across a range of visual inspection tasks.

Similarly, ALFA [42] is a training-free approach that leverages VLMs for zero-shot visual AD, addressing both classification and localization. By introducing a run-time prompt adaptation strategy and a fine-grained aligner, ALFA generates adaptive, informative prompts for each image and achieves precise pixel-level anomaly localization without any task-specific retraining. This method significantly improves detection performance and flexibility over previous approaches, especially in scenarios with limited or no annotated data.

The approach of Xu et al. [43] is another example that demonstrates that a VLM can be customized for anomaly inspection entirely through prompting. By feeding model expert knowledge in the form of task instructions, class-specific context, normalcy rules, and reference normal images, they significantly improved defect detection accuracy and interpretability without model retraining. In a similar approach, Schiele et al. [44] report that one-shot prompting of GPT-4V achieves an image-level anomaly detection F1-score of 92 % on the MVTec-AD benchmark.

In the domain of time series, Russell-Gilbert et al. present RAAD-LLM [45], which couples a frozen pretrained LLM with a RAG module. The approach leverages domain-specific knowledge for AD in predictive maintenance tasks on multivariate sensor data in an extrusion process. RAAD-LLM utilizes reference statistics from a knowledge base and integrates a dynamic updating module to continuously refine what is considered normal behaviour.

In summary, existing FM-based approaches demonstrate expressive performance in finding outliers in industrial settings, and several frameworks for different data modalities have been proposed. Despite the detection performance, a notable advantage identified in the literature is that leveraging FMs for AD enables users additionally to interact with the system in natural language, allowing for deeper investigation and understanding of the root cause. Unlike conventional AD systems that provide only fixed outputs, recent implementations have shown that users can query the model for clarification or request additional details about a detected anomaly, with the model responding in a context-aware and human-readable manner. FMs thus open new opportunities for human-machine collaboration in industrial contexts.

Nevertheless, the literature review also reveals that most existing solutions rely on task-specific model fine-tuning or complex hybrid pipelines, which require significant engineering expertise for initialization and ongoing adjustment. Furthermore, although existing frameworks often support interactive discussions, in most cases it remains challenging to inject new domain knowledge or update the model's expertise at inference time without additional training or engineering modifications. Most approaches either depend on the availability of annotated datasets or existing process knowledge at the system's initialization phase. However, these prerequisites do not accurately reflect the practical challenges of AD in a dynamic industrial setting, where true anomalies are rare, process knowledge is typically acquired incrementally during the ramp-up phase, and anomaly definitions are highly specific to each use case. For example, in computer vision applications, a sudden and subtle change in the colour of a product surface may occur for the first time during a batch change. In such cases, a pretrained FM that lacks use case-specific context might still classify this instance as normal or flawless, as training data often consists of more visible anomalies such as scratches. Additionally, domain experts were not able to initialize the system with this information, since neither the knowledge of existence of this anomaly nor relevant data were available. Consequently, it is evident that enabling domain experts to easily implement and refine anomaly definitions according to their specific requirements is of critical importance.

Therefore, the limitations of most existing frameworks lie in insufficiently addressing the demands of dynamic production environments, where both high outlier detection performance is required and new knowledge and definitions must be incorporated quickly and flexibly. There remains a clear need for frameworks that are designed from the perspective of the domain user, not the data scientist. In practice, the lack of effective human-AI collaboration presents a significant barrier to successfully deploy LLM applications [13].

PB-IAD aims to fill this gap by its pure ICL paradigm. It operates entirely through prompt engineering, does not necessarily require data, retraining or model adaptation for initialization, and can be flexibly adjusted to new anomaly definitions solely by changing the prompt template. This makes it especially well-suited for data-sparse ramp-up situations, where rapid deployment and adaptation are critical and labelled defect data may be unavailable. The framework is intentionally designed to be lightweight rather than complexly engineered, thereby shifting the focus of the quality control system to the domain users, such as factory operators, quality engineers or process experts. The approach enables these users to initialize and adapt the system using natural language instructions alone, without any data science expertise.

## PB-IAD Prompt-based Industrial Anomaly Detection Framework

The proposed framework PB-IAD is designed to leverage FMs for detecting anomalies in industrial data, including images, time series, tabular, and acoustic features, through prompt engineering. It is intentionally developed to be implementable by domain users without any data science or programming background. By leveraging prompt-based instructions in natural language, the framework allows domain experts to directly encode their tacit knowledge and quality criteria into the system, making automated anomaly detection easily adaptable and user-centric. The framework is additionally designed for dynamic industrial environments where data is unavailable and conditions are not stable, as it operates with zero-shot or only few-shot examples. This enables rapid deployment during the ramp-up phase of a new product or production line. As shown in Fig. 3, the architecture of PB-IAD consists of three main components.

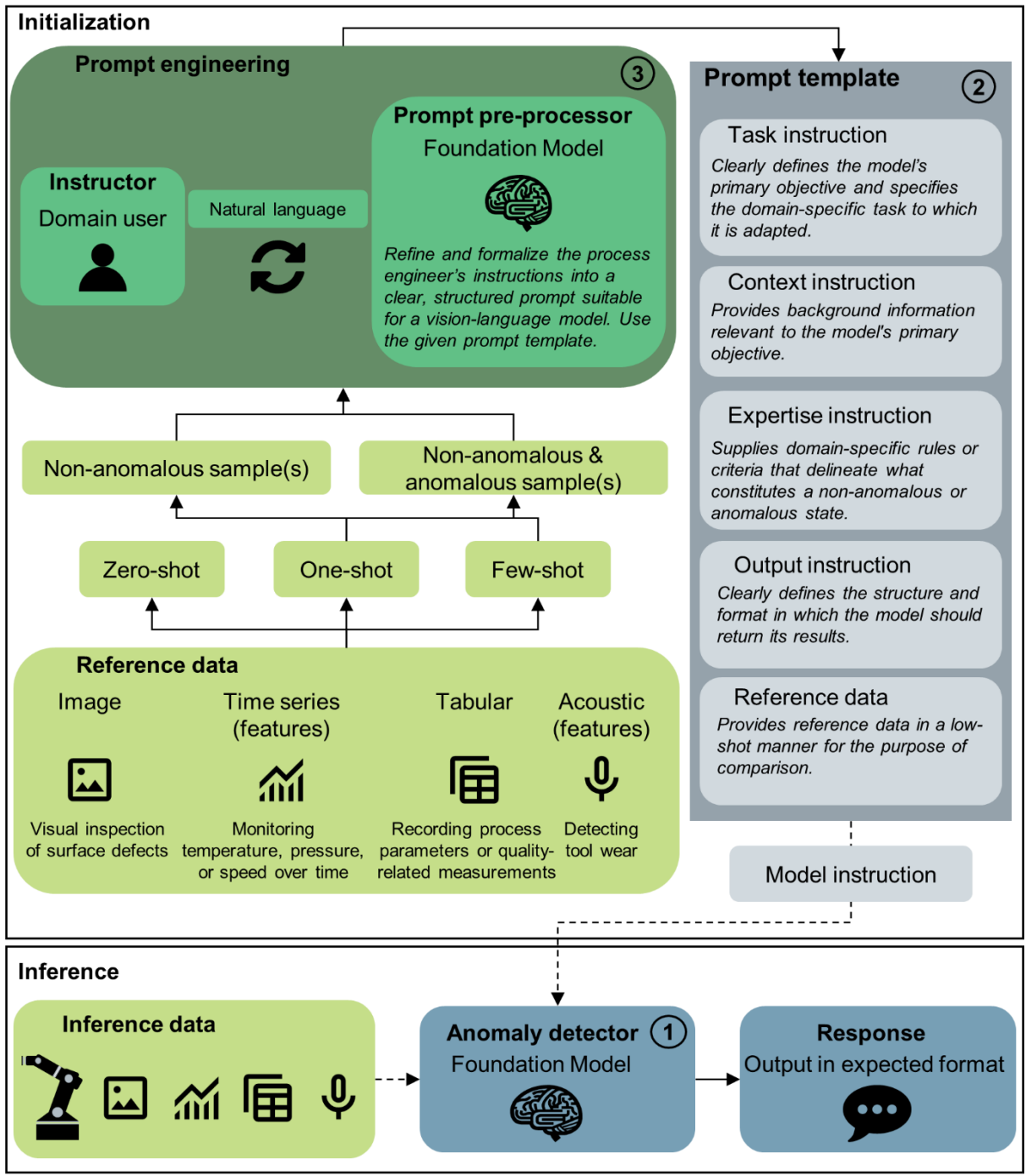

**Fig. 3.** Architecture of the Prompt-based Industrial Anomaly Detection framework.

① **Anomaly detector**: The framework is centred upon a FM that functions as the anomaly detector. The architecture leverages the perception capabilities of the FM acquired though the extensive pretraining. Depending on the availability of computing resources and data security considerations, the slot can be filled variably with either commercial models like GPT4 [20] or locally hosted open-source models like DeepSeek [46].

② **Prompt template**: The prompt template instructs the general FM toward the task of AD and customizes it to the specific use case. As a result, it enables the ICL approach, where no model weights are updated and no retraining is required. The template is composed of four prompt-based and one data-driven section. These sections function as modular components, with their content being populated according to the depth of information available, as depicted in Fig. 4. The definition of the sections is described as follows:

- Task instruction (Ti): The task instruction clearly defines the model's primary objective and specifies the domain-specific task to which it is adapted.
- Context instruction (Ci): The context instruction is intended to provide background information relevant to the model's primary objective. This information should facilitate the model's comprehension of the broader context of the task. It includes general knowledge and common-sense information, yet it is more narrowly focused than the task instruction. The section is designed to be populated with context that is accessible to users without extensive expertise in the specific use case.
- Expertise instruction (Ei): The objective of the expertise instruction is to enrich the prompt with detailed information by domain experts. The section encompasses a range of domain-specific and in-depth rules or criteria that delineate what constitutes a non-anomalous or anomalous state. This section is intended to be populated with context that is accessible to process experts.
- Output instruction (Oi): The output instruction clearly defines the structure and format in which the model should return its results. In contrast to the task instruction, which tells the model what to do, the output instruction specifies how to respond. Techniques such as structured output generation can be employed to ensure the model reliably generates schema-compliant outputs [47].
- Reference data: If available, reference data may be provided within the template in a low-shot manner. This further supplies the model with additional information to improve its semantic instruction to accomplish the AD task.

The prompt template can be populated with information based on the availability of expertise and data. Even with only the task and output instructions, the FM's pretraining enables it to leverage general knowledge to perform an AD task. Furthermore, since the sections function as modular components, the template can be populated progressively and iteratively. As new insights emerge in a ramp-up scenario, they can be incorporated into the template accordingly.

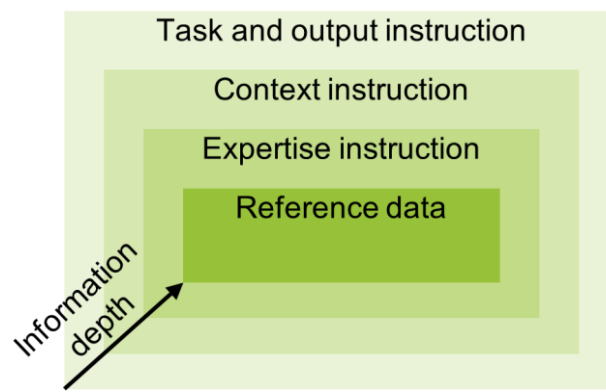

**Fig. 4.** The anomaly detector is instructed by five sections within a prompt template. These sections are filled progressively based on the available depth of information, with deeper layers providing increasingly detailed and specific guidance.

③ **Prompt engineering**

The transition from domain expertise to a usable prompt is managed through a prompt engineering process. Prompt engineering is a distinct module within the framework, defined as the technique of iteratively refining the prompt to enhance its effectiveness [34]. This module incorporates a dedicated FM that functions as a prompt pre-processor, transforming the user's natural-language instructions into a clear, structured prompt aligned with the prompt template. The process is designed to be iterative, allowing domain experts to review and approve refinements before finalizing the prompt for use in the anomaly detection task. This collaborative setup enables domain experts to translate their insights into actionable instructions, even without prior expertise in prompt engineering. General prompt engineering guidelines, for example from OpenAI [48], Anthropic [49] or published best practices [34], may be supplied to the pre-processor as additional context. In scenarios where the prompt template is filled iteratively, the pre-processor also takes the current state of the prompt as context and appends only the newly provided expertise. An exemplary instruction prompt of the pre-processor is depicted in Fig. 5.

> Refine and formalize the process engineer's instructions into a clear, structured prompt suitable for a vision-language model acting as an anomaly detector in manufacturing domain.
>
> Use the given prompt template: <prompt template>. Consider the prompt engineering guidelines: <prompt engineering guidelines>. Adjust the current instruction prompt, if available: <current instruction prompt>.

**Fig. 5.** Exemplary instruction of the prompt pre-processor.

**Input data**

In the context of manufacturing, data is generated in a variety of forms and modalities, depending on the specific monitoring task [50]. Image data, for instance, is frequently utilized in visual inspection tasks to identify surface defects or to directly assess the condition of manufactured products. Time series data is utilized to capture variables such as temperature or speed over time, thereby enabling the monitoring and control of critical process parameters, which directly influence the quality characteristics of the product. Tabular data includes information such as process parameters, quality metrics, or supplier data, which are typically obtained from Manufacturing Execution Systems (MES). Conversely, acoustic data is frequently employed in predictive maintenance tasks, encompassing the identification or prediction of tool wear that has potential to impact product quality or result in production disruptions. The objective of the design of the proposed framework is to leverage the multimodal capabilities of the FM to detect anomalies across all necessary data modalities in manufacturing. At the inference stage, the data to be classified is submitted to the anomaly detector as part of the user prompt. Depending on the availability of data, the instruction prompt may optionally include zero-shot, one-shot, or few-shot reference samples, such as examples of non-anomalous or anomalous states, as described in the prompt template section.

In summary, the framework is built upon the essential requirements for implementing a quality control system in a dynamic production environment. The approach is characterised by its emphasis on data sparsity, multimodal capabilities, and domain user centricity. It utilizes ICL, enabling rapid deployment and adaptation solely through semantic instruction and low-shot reference samples. Consequently, the approach obviates the necessity for retraining or massive labelled data and enables domain experts with no prior data science experience to configure and adapt the system independently.

# Experimental setup and evaluation methodology

In order to provide a comprehensive evaluation of the proposed framework in a range of industrial scenarios, three distinct experimental setups were designed, each of which reflects a different manufacturing condition. In all experiments, GPT-4.1 (utilizing gpt-4.1-2025-04-14 snapshot) was employed as the anomaly detection model, while GPT-4o (chatgpt-4o-latest snapshot) functioned as the prompt pre-processing engine. Furthermore, the task and output instruction, as illustrated in Fig. 6, remained uniform throughout all configurations to ensure consistency.

> ## TASK INSTRUCTION:
> You are an anomaly detection assistant.
> Your task is to determine whether the TEST-SAMPLE is anomalous or non-anomalous.
> - 0 = Non-anomalous
> - 1 = anomalous

> ## OUTPUT INSTRUCTION:
> Respond only with a JSON object like this:
> {"Classification": <0 or 1>, "Reasoning": <explanation>}

**Fig. 6.** Task and output instruction of the anomaly detector over all scenarios.

The task instruction provides the anomaly detector with the general instruction of its objective. The output instruction expects a JSON format including the classification result, anomalous (1) or non-anomalous (0) state, and a reasoning component that explains the model's decision. This reasoning component serves to reveal potential weaknesses or hallucinations in the model's decision-making process. Conversely, as utilized in various publications, it is not intended for root cause analysis of the cause of the anomaly, but rather for enhancing the reliability and transparency of the quality control system itself. This approach can be compared to the visualization of threshold deviations in statistical algorithms, which similarly provide insight into the reasons why the system classified a particular instance as anomalous.

## Evaluation

The performance of the three experiments is primarily evaluated using Precision, Recall, and F1- score. A low precision value is indicative of a high false positive rate, which, in a manufacturing context, results in increased scrap costs, as non-anomalous products are incorrectly classified as defective. Conversely, the recall metric is indicative of the false negative rate, where anomalous products are categorized as non-anomalous. Consequently, this can lead to safety-critical concerns, as defective products may incorrectly pass the quality control stage. Finally, the F1-score provides a balanced measure that captures the trade-off between precision and recall.

To systematically evaluate the influence of the information depth provided to the model, an ablation study is conducted with increasing levels of prompt enrichment:

- Ti_Oi: Baseline setup with only the task and output instruction provided
- Ti_Oi_Ci: Adds the context instruction to the baseline configuration
- Ti_Oi_Ci_Ei: Further incorporates expertise-related guidance into the prompt
- Ti_Oi_Ci_Ei_Rd: Additionally includes reference data

When no reference data is provided, the task is classified as zero-shot anomaly detection. The provision of non-anomalous instances defines the experiment as one-shot or few-shot anomaly detection tasks, depended on the number of samples. When both non-anomalous and anomalous examples are included, the scenario is considered as a binary classification task.

In addition to the ablation study, the effectiveness of the framework is evaluated by benchmarking it against ML-based anomaly detection algorithms. The evaluation is specifically focused on scenarios characterized by data sparsity. The ML models are iteratively trained on increasing amounts of data to identify the point at which they begin to outperform the proposed framework. This approach facilitates the evaluation of whether the ICL-based framework is more effectively designed for rapid and early stage integration in dynamic industrial environments, where data is limited and conditions are subject to frequent change.

## Scenario 1: Visual inspection in stable manufacturing conditions – MVTec subset cable

The first experiment is designed to simulate a stable manufacturing environment, characterized by consistent conditions, minimal product variation, and a fixed data acquisition setup without any modifications. For this purpose, a subset of the open-source and widely adopted MVTec Anomaly Detection dataset [51] is employed. This benchmark dataset comprises over 5,000 high-resolution colour images of various objects. Its low variability and uniform lighting conditions [52] make it particularly suitable for modelling stable operational scenarios. Specifically, the "Cable" category is selected for this experiment. The dataset contains a total of 224 images labelled as "good" in the training set and 58 "good" images in the test set. The dataset also comprises 92 anomalous images distributed across eight distinct defect types. In accordance with the zero- to low-shot objective of the proposed framework, only the first image from the training set is retained as a one-shot reference. All remaining training images have been omitted.

Two experiments are conducted within this scenario:

- Zero-shot anomaly detection: No reference image is provided
- One-shot anomaly detection: A single OK reference image is provided

The prompts utilized for the context and expertise instruction for this scenario are provided in the Appendix A in the supplementary material (Fig. 12 and Fig. 13). The context instruction comprises all information that a non-expert could use to guide the model, such as what is visible in the images, variations in lighting or background, and circumstances in cable structure or insulation colours. It is important to note that these details do not necessitate in-depth process knowledge. In contrast, the expert instruction section provides a detailed explanation of the process of distinguishing anomalous from non-anomalous states. Preliminary tests revealed that it is crucial to instruct the model that not all anomaly criteria need to be met for a sample to be considered anomalous, instead a single matching criterion is sufficient. In the absence of this clarification, the model wants the sample to match all anomaly descriptions to be classified as anomalous. Table 1 summarizes the ablation study of this scenario and provides the model's outputs for better comprehension.

**Table 1.** Experimental setup of Scenario 1, with an example query of an anomalous image showing poked insulation. Green highlights indicate correct outputs of the anomaly detector, while red highlights indicate incorrect outputs.

| Input | | | Output | |
|---|---|---|---|---|
| Instruction | Reference sample | Query | 0/1 | Result |
| Ti, Oi | | 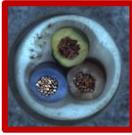 | 0 | The image shows a cross-section of a multi-core electrical cable, with three distinct wires (with blue, yellow/green, and brown/grey insulation) and copper strands inside. The wires appear intact, with no visible damage, burning, corrosion, or fraying. The insulation also appears to be complete and unbroken, and the distribution of wires is typical for such cables. There is no sign of anomaly present. |
| Ti, Oi, Ci | | 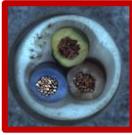 | 1 | The cable cross-section shows significant structural defects in the outer insulating sheath, which is cracked and missing substantial material, exposing the inner insulation of the conductors. This compromises the cable's integrity and could lead to electrical and environmental hazards. All three conductors and their respective insulation colors are present and appear normal, but the main jacket damage is a clear anomaly. |
| Ti, Oi, Ci, Ei | | 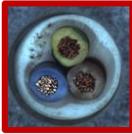 | 1 | The outer white insulating sheath of the cable shows visible damage with multiple cuts or notches on the left side, indicating a breach. This constitutes a 'Cut Outer Insulation' anomaly. All three inner conductors and their copper strands appear intact, but the outer jacket damage is sufficient to classify this as anomalous. |
| Ti, Oi | 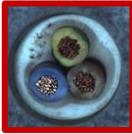 | 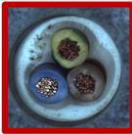 | 0 | The TEST-SAMPLE image shows three copper wires encased in colored insulating material, similar in configuration, coloration, and structure to the NON-ANOMALOUS-SAMPLE. There are no visible deviations, damage, or irregularities that indicate an anomaly, such as missing or damaged wires, discoloration, or foreign objects. |
| Ti, Oi, Ci | 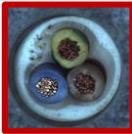 | 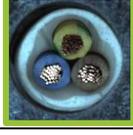 | 1 | The TEST-SAMPLE image shows significant damage to the outer insulating sheath, with large irregular gaps and voids visible primarily at the top left area of the cross-section. This is in contrast to the NON-ANOMALOUS-SAMPLE image, where the outer sheath is smooth and continuous. The presence of such structural defects in the TEST-SAMPLE could potentially compromise cable safety and function, indicating an anomaly. |
| Ti, Oi, Ci, Ei | 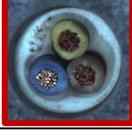 | 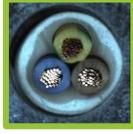 | 1 | In the TEST-SAMPLE image, there is observable damage to the outer white sheath, with multiple cracks and breaks in the insulation near the top left section. This constitutes a 'cut outer insulation' anomaly as defined in the criteria. All conductor colors and copper strands appear correct and in place otherwise, but any single anomaly is grounds for a positive detection. The NON-ANOMALOUS-SAMPLE shows no such damage to the outer sheath, providing a good reference for comparison. |

## Scenario 2: Visual inspection in dynamic manufacturing conditions – Stripped wires

The second scenario consists of a custom-designed collection of images illustrating stripped conductors captured under a wide range of conditions [53]. These include variations in conductor colour, lighting, viewing angle, insulation quality, image resolution, and stripping length, which collectively reflect a dynamic production environment. Furthermore, in comparison with the dataset of Scenario 1, the anomalies are more subtle and fine-grained, thereby more accurately reflecting the challenges encountered in quality control. The dataset consists of both normal samples and two types of anomalies: pulled strands and cut strands. One sample of each quality class is maintained as a reference sample. As demonstrated in Fig. 7, it is important to note that solely the red-white insulation type is employed as the reference; the test samples encompass a more diverse range of conditions. In total three experiments are conducted within this scenario:

- Zero-shot anomaly detection: No reference image is provided.
- One-shot anomaly detection: A single OK reference image is provided.
- One-shot binary classification: A single image of each quality class is provided.

The context and expert instructions, are provided in Appendix A in the supplementary material (Fig. 14 and Fig. 15).

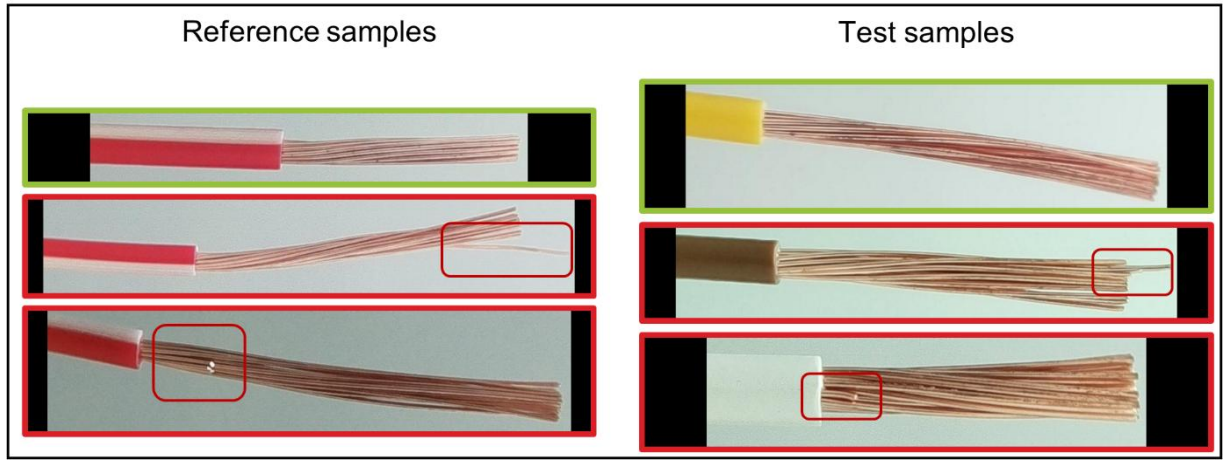

**Fig. 7.** Reference samples, one of each quality class and explementary test samples of Scenario 2. The red squares highlight the more subtle anomalies compared to the MVTec Dataset.

### Scenario 3: Discrete time series features – Crimp force curve

In the third experimental scenario, discrete time series features serve as the data modality. The dataset is drawn from the Crimp Force Curve Dataset [54], which contains annotated force-displacement curves obtained from a semi-automatic crimping machine. Each time series consists of 500 data points, and for this particular experiment, a randomly chosen subset of 150 curve samples was selected, all of which corresponded to wires with a cross-section of 0.50 mm². Of the 150 samples examined, 50 are labelled as non-anomalous, while the remaining 100 are anomalous, equally distributed between two defect types: missing strands and crimped insulation. The experiment is structured as a few-shot learning task, with three instances from the non-anomalous class provided as reference samples. This configuration is consistent with industrial practices, where the presence of variability and noise in time series data necessitates the utilisation of multiple reference examples to ensure the reliability of analysis. Furthermore, in practical applications, the full force curve is rarely used directly for quality control. Instead, selected features such as the area under the curve at specific intervals are typically extracted and analysed, as described in common industrial monitoring approaches [55,56]. For this particular scenario, the slope between datapoint 150 and 190 and the area under the curve, which reflects the applied force, between data point 250 and 300 were selected as quality features.

Although feature computation was carried out externally in this experiment, prior evaluations have demonstrated that the model is capable of autonomously calculating these features when coding capabilities are enabled or when access to external tools is provided. Based on the given instructions, the model is capable of extracting the two statistical characteristics from the raw 500-point force curves independently. However, due to constraints related to inference time, API costs for external tool usage, and the absence of fully integrated coding functionality, the features were pre-computed and directly supplied as model input. The configuration of Scenario 3 is illustrated in Fig. 8. Additionally, the context and expertise instruction are provided in Appendix A in the supplementary material (Fig. 16 and Fig. 17).

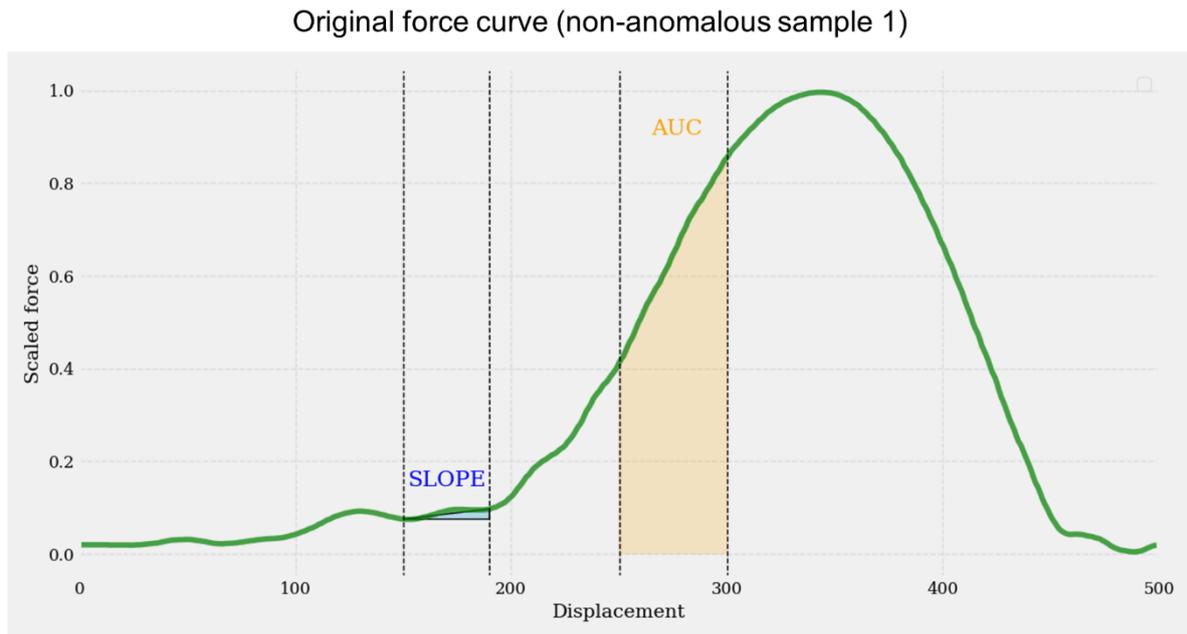

**Fig. 8.** In Scenario 3 the anomaly detector is provided with precalculated features of a time series as few-shot reference samples.

## Ablation study

The ablation study evaluates the impact of the information depth of the model instruction. The results of the study are summarised in Table 2, which includes all three experimental scenarios. In addition to the performance metrics of precision, recall, and F1-score, the table also includes the number of input tokens and the average number of output tokens.

**Table 2.** Results of the ablation study of all three scenarios.

| Experiment | Information depth of prompt | | | | Evaluation metrics | | | | |
|---|---|---|---|---|---|---|---|---|---|
| | Ti | Oi | Ci | Ei | Precision | Recall | F1-Score | Input Tokens | Ø Output Tokens |
| **Scenario 1: Visual inspection in stable manufacturing conditions – MVTec subset cable** | | | | | | | | | |
| Zero-shot anomaly detection | ✓ | ✓ | | | 97,1 % | 73,9 % | 84,0 % | 886 | 86 |
| | ✓ | ✓ | ✓ | | 97,5 % | 83,7 % | 90,1 % | 1080 | 99 |
| | ✓ | ✓ | ✓ | ✓ | 98,8 % | 88,0 % | 93,1 % | 1470 | 101 |
| One-shot anomaly detection | ✓ | ✓ | | | 95,4 % | 89,1 % | 92,1 % | 1669 | 100 |
| | ✓ | ✓ | ✓ | | 95,6 % | 93,5 % | 94,5 % | 1863 | 108 |
| | ✓ | ✓ | ✓ | ✓ | 95,7 % | 95,7 % | 95,7 % | 2253 | 108 |
| **Scenario 2: Visual inspection in dynamic manufacturing conditions – Stripped wires** | | | | | | | | | |
| Zero-shot anomaly detection | ✓ | ✓ | | | 100.0 % | 46.1 % | 63.1 % | 713 | 76 |
| | ✓ | ✓ | ✓ | | 100.0 % | 63.5 % | 77.7 % | 855 | 69 |
| | ✓ | ✓ | ✓ | ✓ | 100.0 % | 78.4 % | 87.9 % | 1130 | 68 |
| One-shot anomaly detection | ✓ | ✓ | | | 84.2 % | 83.2 % | 83.7 % | 1316 | 90 |
| | ✓ | ✓ | ✓ | | 98.6 % | 82.0 % | 89.5 % | 1458 | 87 |
| | ✓ | ✓ | ✓ | ✓ | 98.6 % | 84.4 % | 91.0 % | 1733 | 84 |
| One-shot binary classification | ✓ | ✓ | | | 98.8 % | 48.5 % | 65.1 % | 2522 | 89 |
| | ✓ | ✓ | ✓ | | 100.0 % | 61.1 % | 75.8 % | 2664 | 92 |
| | ✓ | ✓ | ✓ | ✓ | 100.0 % | 74.3 % | 85.2 % | 2939 | 82 |
| **Scenario 3: Discrete time series features – Crimp force curve features** | | | | | | | | | |
| Few-shot anomaly detection | ✓ | ✓ | | | 100.0 % | 76.0 % | 86.4 % | 324 | 124 |
| | ✓ | ✓ | ✓ | | 100.0 % | 82.0 % | 90.1 % | 472 | 135 |
| | ✓ | ✓ | ✓ | ✓ | 100.0 % | 92.0 % | 95.8 % | 763 | 121 |

The findings from Scenario 1 illustrate that the incorporation of supplementary context leads to a substantial enhancement in the efficacy of AD. In the zero-shot baseline condition, where no supplementary context or expertise instruction was provided, the model achieved a precision of 97.1 %, a recall of 73.9 %, and an F1-score of 84.0 %. Given that the model operated solely based on the task instruction and its general pre-trained knowledge, this outcome is already notable. The incorporation of contextual instruction resulted in a substantial enhancement in performance, increasing the F1-score to 90.1 %. This performance was further improved to 93.1 % through the integration of expertise instruction. These findings emphasise the efficacy of an ICL approach that relies exclusively on semantic guidance. In the one-shot scenario, where the model was provided with one single non-anomalous reference image, the baseline condition yielded a lower precision of 95.4 % but a higher recall of 89.1 %, resulting in a F1-score of 92.1 %. In this particular instance, the model tended to generate a higher number of false positives, owing to the misclassification of certain non-anomalous images as anomalous. As illustrated in Table 3, the decrease in precision can be attributed to the direct comparison of the model with the non-anomalous sample, resulting in a more stringent evaluation.

**Table 3.** Explementary misclassified instance in Scenario 1, in which a one-shot setup performs worse than a zero-shot configuration.

| | Input | | Output | |
|---|---|---|---|---|
| Instruction | Reference sample | Query | 0/1 | Result |
| Ti, Oi, Ci, Ei | | 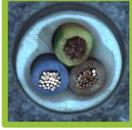 | 0 | All three conductors (yellow-green, blue, brown/grey) are present and the insulation around each is intact with no visible cuts, cracks, or deformations. The copper strands are uniformly packed, with none visibly bent, protruding, or missing. The outer sheath fully encases all conductors and shows no signs of damage. No anomalies such as cut insulation, missing wires, cable swap, or poking copper strands are evident. |
| Ti, Oi, Ci, Ei | 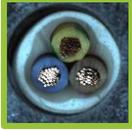 | 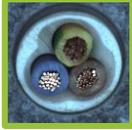 | 1 | In the TEST-SAMPLE image, several copper strands within the yellow-green conductor appear bent and misaligned rather than uniformly packed. The strands at the periphery of the bundle are not sitting flat inside the insulation but protrude slightly, suggesting a bent wire anomaly. In contrast, the NON-ANOMALOUS-SAMPLE shows all copper strands neatly packed and fully within the insulation boundaries. Therefore, the TEST-SAMPLE is classified as anomalous due to bent/misaligned copper strands. |

Overall, the model achieved better in the one-shot setting compared to the zero-shot configuration. Following the integration of contextual instruction, the F1-score demonstrated a notable enhancement, reaching 94.5 %. The incorporation of a combination of contextual and expert instruction further augmented performance to a F1-score of 95.7 %. Consequently, the provision of more comprehensive information in the prompt has resulted in enhanced outcomes. A comparison between the zero-shot and one-shot settings demonstrates that the incorporation of a reference image enhances the model's capacity to detect anomalies. This effect is especially pronounced in the baseline setting, where the F1-score increased from 84.0 % to 92.1 %. However, when detailed instructions are provided, the performance discrepancy between zero-shot and one-shot narrows considerably, from 93.1 % to 95.7 %, a difference of only 2.6 %. This finding indicates that, while reference images continue to improve performance, their relative importance diminishes when the model is instructed by precisely crafted and detailed prompts.

The experimental results for Scenario 2 illustrate how varying degrees of semantic instruction affect the performance of anomaly detection and binary classification in dynamic conditions. This scenario is characterised by frequent changes in production system settings and the presence of fine-grained anomalies. In the zero-shot anomaly detection experiment, the baseline condition achieved a precision of 100 % but a low recall of 46.1 %, resulting in an F1-score of 63.1 %. The incorporation of contextual instruction resulted in a substantial enhancement in recall, achieving 63.5 %, thereby a higher F1-score of 77.7 %. The incorporation of expert instruction resulted in a further increase in recall to 78.4 %, yielding an F1-score of 87.9 %. The findings demonstrate again that the incorporation of contextual information enhances the performance of AD, particularly by improving recall and reducing false negatives. In the one-shot anomaly detection experiment, where a single non-anomalous reference image was provided, the baseline condition achieved balanced precision and recall values of 84.2 % and 83.2 %, respectively, resulting in an F1-score of 83.7 %. The incorporation of contextual instruction resulted in a substantial enhancement in precision, reaching 98.6 %, while recall exhibited a modest decline to 82.0 %, with a total F1-score of 89.5 %. The incorporation of expert instruction led to a significant enhancement in recall, achieving a maximum of 84.4 %, which resulted in the highest F1-score in this particular scenario, reaching 91.0 %. It is noteworthy that in both scenarios, precision tends to decrease in the one-shot setting compared to zero-shot, while recall shows significant improvement. Moreover, as also demonstrated in Scenario 1, the recall improvement from contextual enrichment is less pronounced in the one-shot experiment than in the zero-shot configuration. This can be attributed to the significantly lower baseline performance in the zero-shot setting, which leaves potential for enhancement. The general performance of Scenario 2 is lower than in Scenario 1, which is indicative of the more challenging nature of the dataset. The higher degree of variability in conditions, together with the existence of more subtle anomalies, poses a higher level of difficulty in the AD task. In the one-shot binary classification task, the baseline experiment achieved a high level of precision (98.8 %) but a low level of recall (48.5 %), resulting in an F1-score of 65.1 %. The incorporation of contextual instruction led to a significant enhancement in recall, with 61.1 %. This development subsequently resulted in an F1-score of 75.8 %. The incorporation of expert instruction resulted in the highest recall of 74.3 %, corresponding to the best F1-score of 85.2 % for this classification task. Despite these enhancements, the recall in the binary classification setting remains lower than in the anomaly detection experiments. Similarly to the findings in Table 3, this suggests that the explicit provision of anomalous examples may potentially compromise the model's generalizability or to trigger a stricter comparison. This interpretation is supported by the analysis of model reasoning demonstrated in Table 4.

**Table 4.** Exemplary misclassified instance in Scenario 2, in which a one-shot binary classification setup performs worse than a one-shot anomaly detection configuration.

| Instruction | Input | | Output | |
|---|---|---|---|---|
| | Reference sample | Query | 0/1 | Result |
| Ti, Oi, Ci, Ei | 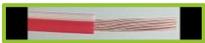 | 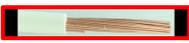 | 1 | The TEST-SAMPLE shows a wire with one copper strand sticking out and separated from the rest, which is not seen in the NON-ANOMALOUS-SAMPLE. In the NON-ANOMALOUS-SAMPLE, all copper strands are neatly twisted together without any stray wires. The presence of the stray strand in the TEST-SAMPLE indicates a defect or anomaly. |
| Ti, Oi, Ci, Ei | 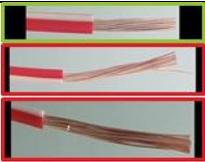 | 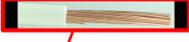 | 0 | The TEST-SAMPLE image shows a wire with copper strands that are mostly uniform and straight, similar to the NON-ANOMALOUS-SAMPLE. In contrast, the ANOMALOUS-SAMPLES have strands that are significantly frayed, separated, or contain visible defects. There is a slight separation in one strand in the TEST-SAMPLE, but it is not as pronounced as in the anomalous examples, so it is classified as non-anomalous. |

In this particular instance, the model accurately identified a sample as anomalous during the anomaly detection task, yet misclassified the same sample as non-anomalous in the binary classification task. The explanation provided was that the defect was "not as pronounced as in the anomalous examples". In general, the experiments demonstrate that the incorporation of contextual and expert information into the semantic model instruction leads to a consistent enhancement in F1-scores across a range of AD scenarios. Moreover, a comparison of zero-shot and one-shot performance reveals that the one-shot configuration offers a substantial advantage in the baseline setting (83.7 % compared to 63.1 %). However, it is important to note that this benefit is diminished when the model is supplied with complete contextual information. Moreover, the incorporation of reference samples may enable the model to concentrate on this particular example, potentially at the expense of its generalisability as the results of the one-shot binary classification (85.2 %) compared to the one-shot AD (91.0 %) showed.

The third scenario explores the use of discrete time series features derived from crimp force curves. The baseline experiment, which does not incorporate any additional context, achieved an F1-score of 86.4 %. The implementation of contextual instruction resulted in a better performance, with an F1-score of 90.1 %. The incorporation of expertise instruction resulted in a further enhancement of performance of 95.8 %. This outcome is comparable with that of prior research studies [57] that employed ML methodologies for the purpose of crimp force monitoring. Despite the absence of direct comparability, owing to discrepancies in dataset diversity and complexity, the findings underscore the efficacy of FM-based AD in time series domain. As illustrated by the reasoning example in Table 5, the model, when operated without supplementary context, is still capable of accurately identifying the misalignment in the AUC feature. However, in the absence of a clear point of reference, it is up to the system to determine when a deviation should be considered anomalous. This has been demonstrated to result in a tendency towards inconsistent decision-making.

Table 5. Exemplary misclassified instance in Scenario 3, in which missing contextual information leads the model set its threshold by itself.

| Instruction | Input | | Output | |
|---|---|---|---|---|
| | Reference sample | Query | 0/1 | Result |
| Ti, Oi | 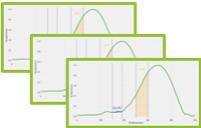 | 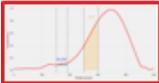 | 0 | The TEST-SAMPLE's SLOPE datapoint (0.0004055) is within the range of the OK samples (0.0002982 to 0.0009065). The AUC datapoint (29.2021) is slightly lower than the OK samples (which range from about 31.19 to 32.41), but the difference is minor (~2 units below the lowest OK sample) and does not indicate a drastic deviation or clear anomaly. Therefore, the TEST-SAMPLE is classified as non-anomalous. |
| Ti, Oi, Ci, Ei | 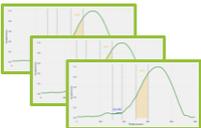 | 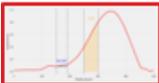 | 1 | The TEST-SAMPLE's slope (0.0004055) in region 150-190 is comparable to reference slopes (ranging from 0.0002982 to 0.0009065), so there is no anomalous steepness. However, the area under the curve (AUC) from 250-300 is 29.20, which is lower than all reference samples (the lowest being 31.19). This reduced AUC indicates the test sample exhibits lower force values in this critical region, matching the anomalous criteria regarding the AUC. |

Conversely, when expert knowledge is provided, specifically delineating those values outside a defined range should be treated as anomalies, the model accurately classifies the example. Preliminary investigations also demonstrated that smaller language models were more prone to hallucinate when comparing numerical values. The GPT-4.1 model utilized in this study did not exhibit such behaviour and produced consistent, and reliable outputs. The overall precision of the model remained at 100 % across all configurations, indicating that no false positives were produced. As more contextual information was incorporated, there was an observed improvement in recall, thus indicating an enhancement in the model's sensitivity to anomalies.

Across the full range of scenarios examined, the experiments demonstrated a consistent improvement in AD performance when additional context was provided through ICL. The more precise instruction the model receives, the better it performs. The incorporation of low-shot examples has been demonstrated to further enhance performance. However, the impact of such examples is less significant than that of clearly written instructions. Overall, from the three examined scenarios, it can be derived, that clear and well-structured semantic instructions, enable a multimodal FM for the downstream task of IAD.

## Comparing performance to other machine learning algorithms

In order to contextualize the performance of the proposed approach, the obtained results from the ablation study are compared to ML-based algorithms, which represent a potential alternative approach for the AD task. The objective is to evaluate whether the PB-IAD framework provides superior quality control performance compared to ML-based methods under dynamic manufacturing conditions. The primary objective of this study is to determine the number of input samples necessary for the ML-based approach to match the performance of PB-IAD.

For the first two scenarios, the PatchCore algorithm [58] is employed as the reference model, selected due to its documented SOTA performance on the MVTec Anomaly Detection dataset. The algorithm functions by creating a memory bank of nominal feature representations from local image patches. Anomalies are subsequently identified by measuring the distance of test patches to their nearest neighbours in this memory bank. To ensure a robust and standardized comparison, the model was configured using hyperparameters previously shown to be effective on the MVTec benchmark dataset. Specifically, the configuration employed a Wide ResNet-50 architecture as the feature extraction backbone, sourcing feature maps from its second and third intermediate layers. A coreset sampling ratio of 0.1 was applied to construct the final memory bank. To simulate a data ramp-up scenario, the algorithm is trained on incrementally larger sets of normal samples from both the MVTec Cable and the custom Stripped Wire datasets, with training set sizes ranging from 5 to 200 images. To establish an optimal decision threshold for the PatchCore model a validation subset comprising 20 % of the samples was partitioned from the original test set. The final performance of PatchCore is subsequently reported on the remaining 80 % hold-out test data.

In Scenario 1, representing the stable production environment with the MVTec Cable dataset, the performance of the PatchCore algorithm demonstrates a clear positive correlation between the number of training samples and detection accuracy. The results are illustrated in Fig. 9, which plots the PatchCore results against the one-shot PB-IAD performance benchmarks. With only 5 training samples, the model achieves an F1-score of 78.33 %. Performance rapidly improves as more data is introduced, crossing the 0.90 F1-score threshold at 30 samples. The results indicate that performance gains begin to saturate as the training set size approaches 100 samples, where the model achieves a peak F1-score of 95.67 %. At this point, the model is supported by strong and balanced precision and recall values of 95.51 % and 95.85 %, respectively. Beyond this, adding more data yields only marginal improvements, suggesting that for a stable production scenario, PatchCore can achieve robust performance with a relatively modest number of training images. The F1-score plot in Fig. 9 shows that PatchCore requires approximately 60 training samples to surpass the baseline PB-IAD configuration's F1-score of 92.1 %. To match the 95.7 % F1-score of the fully-instructed PB-IAD model, the data-driven approach needs around 100 training samples. This demonstrates that even in a stable, low-variance environment where PatchCore is highly effective, the instruction-driven PB-IAD framework achieves superior or equivalent performance with only a single reference image, highlighting its significant data efficiency.

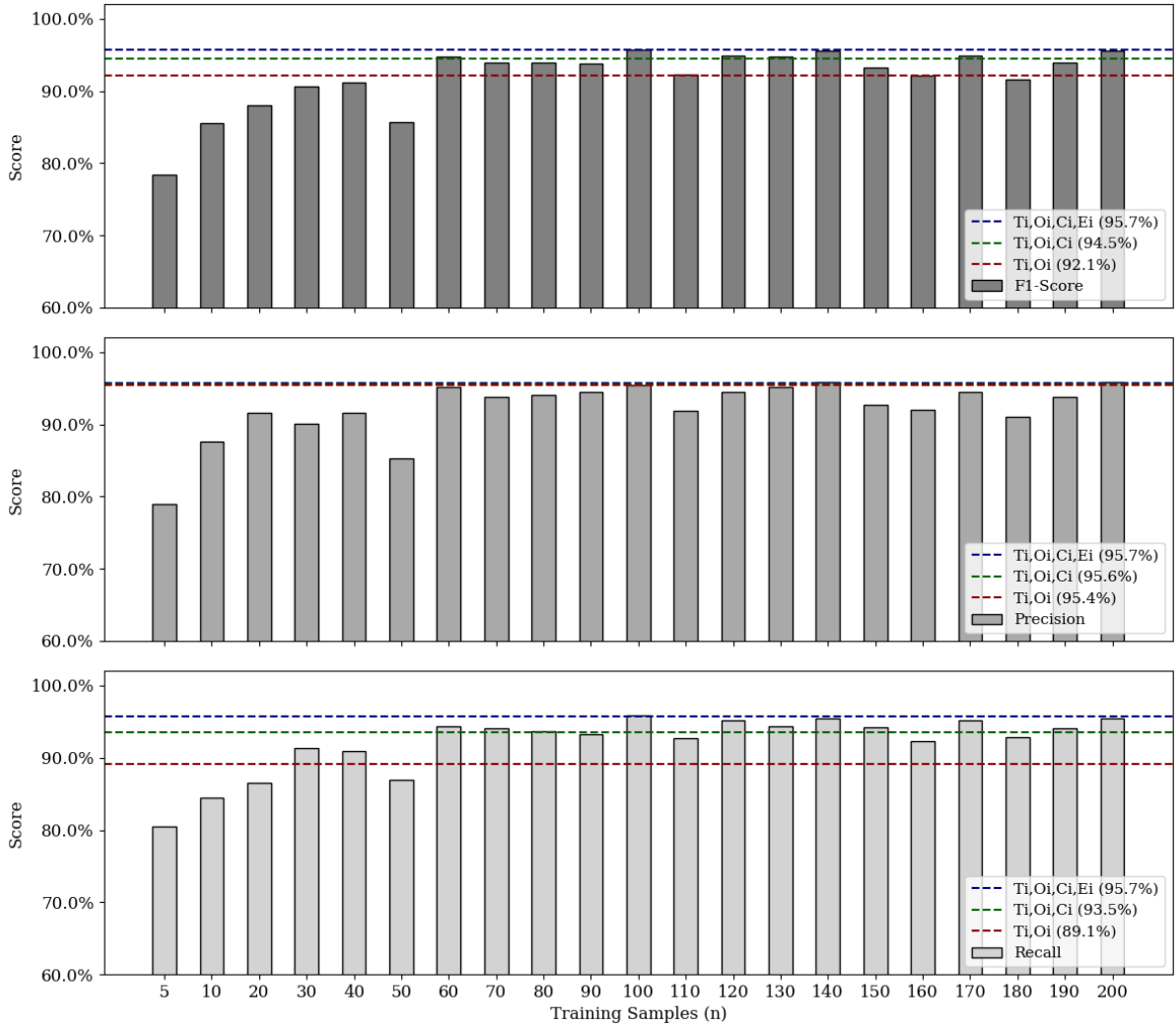

**Fig. 9.** PatchCore results of Scenario 1 benchmarked against PB-IAD.

In contrast, the evaluation for Scenario 2, conducted on the custom Stripped Wire dataset, reveals the challenges posed by a more dynamic and complex scenario. Here, the model's performance is not only lower but also significantly more erratic. Performance starts low with an F1-score of 66.15 %, and unlike the steadier progression observed previously, the learning curve is characterized by high volatility. For instance, at 110 training samples, the model exhibits a significant performance imbalance: while precision reaches a high of 83 %, recall drops sharply to 72.62 %, resulting in a poor F1-score of only 72.31 %. This indicates that the model fails to detect a substantial proportion of actual anomalies, likely due to conservative prediction thresholds. The model surpasses an F1-score of 80 % only after being trained on 100 samples and reaches its peak F1-score of 83.82 % with 190 samples, where its precision and recall achieve a strong balance at 84.67 % and 84.79 %, respectively. The lower overall scores and fluctuating performance underscore the difficulty the data-driven model faces in generalizing from a small sample set when confronted with high variability and subtle defect types. This indicates that a substantially larger training dataset is necessary for PatchCore to reliably model the normal class under such dynamic conditions. This performance disparity is visually summarized in Fig. 10, where the one-shot PB-IAD results are overlaid as horizontal benchmarks. The F1-score plot highlights the gap most clearly. PatchCore requires approximately 190 training samples to match the performance of the most basic PB-IAD configuration. Furthermore, even with 200 samples, the data-driven model's F1-score remains significantly below the levels achieved by PB-IAD when enriched with context (Ci) and expert (Ei) instructions. A similar trend is observed for precision, where PatchCore never approaches the 98.6 % benchmark set by the informed PB-IAD prompts. This illustrates that in the dynamic, high-variance conditions of Scenario 2, the instruction-driven, one-shot PB-IAD approach is not only more data-efficient but also delivers superior and more stable performance compared to a conventional data-driven model.

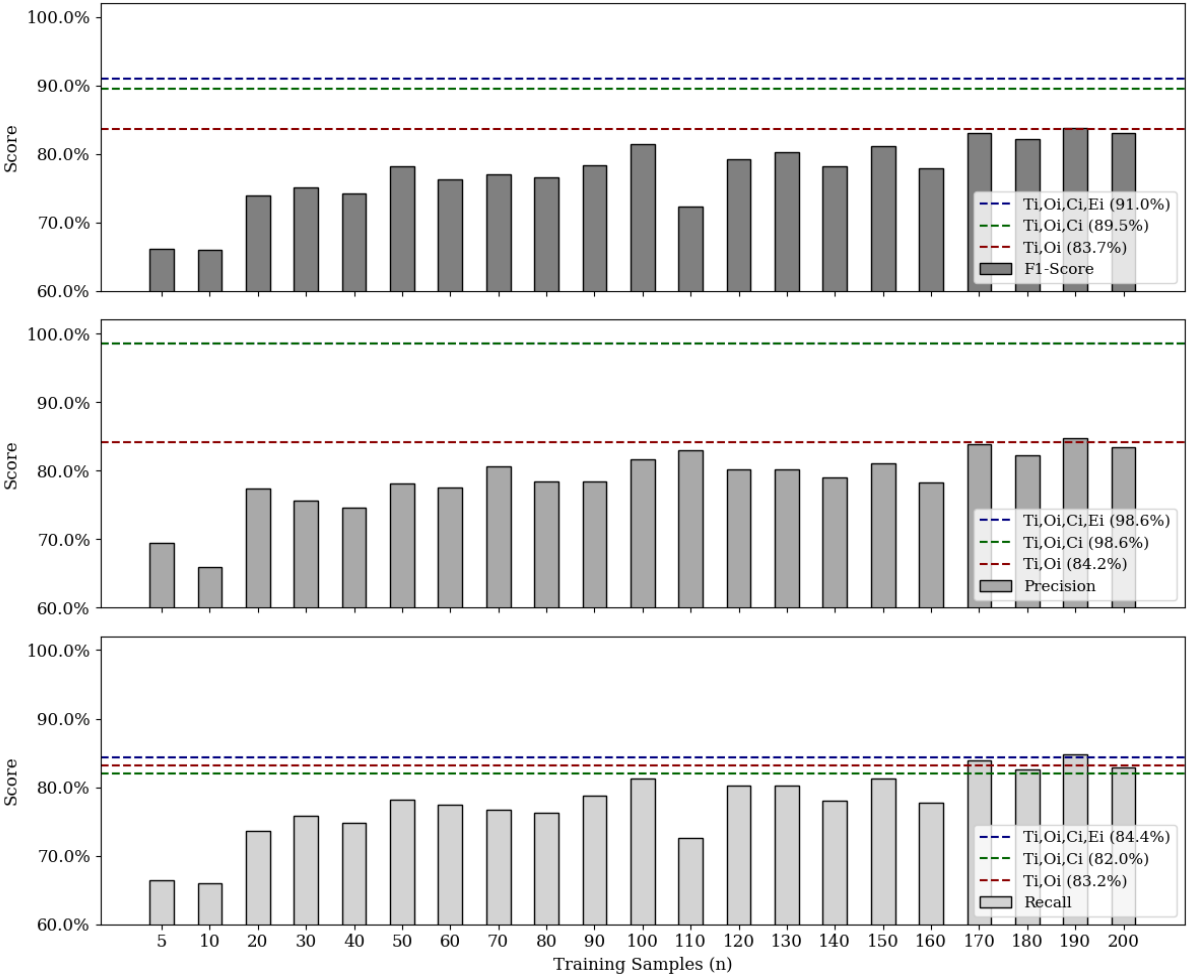

**Fig. 10.** PatchCore results of Scenario 2 benchmarked against PB-IAD.

Taken together, the results of both visual inspection scenarios reveal a key trade-off in the detection of anomalies in manufacturing. In a stable environment (Scenario 1), a data-driven model like PatchCore can achieve high performance, eventually matching the one-shot PB-IAD framework, but only after being trained on a substantial dataset of approximately 100 samples. However, in a dynamic environment marked by high process variability (Scenario 2), the performance of the data-driven model deteriorates significantly, failing to match even the most basic PB-IAD configuration without a large number of training samples. This demonstrates that while classical models are effective, their performance is fragile and highly dependent on data consistency. The PB-IAD framework's ability to codify expert knowledge provides the necessary robustness, making it a fundamentally more data-efficient and reliable solution for production ramp-ups and other data-scarce, dynamic industrial settings.

In the third benchmark scenario, the Isolation Forest (IF) algorithm [59] from scikit-learn library [60] is employed as a reference model. The corresponding results are presented in Fig. 11. The performance of each input sample configuration, ranging from 5 to 200 samples, is reported based on the optimal contamination hyperparameter $C$, which was selected from the discrete search space $C \in \{0.10, 0.15, 0.20, 0.25, 0.30, 0.35, 0.40, 0.45, 0.50\}$.

All other hyperparameters were kept at their default settings. Despite utilising merely three labelled examples, the few-shot PB-IAD scenario attains F1-scores of 86.4%, 90.1%, and 95.8%, depending on the instruction's information depth. In contrast, IF attains comparable performance only after a significantly larger number of samples: approximately 10 samples for the Ti,Oi setup, 50 samples for Ti,Oi,Ci, and around 110 samples for the Ti,Oi,Ci,Ei configuration. It is noteworthy that, even with 200 samples, the performance of IF does not demonstrate a substantial enhancement over PB-IAD. The recall metric is the sole exception to this, as it becomes superior from approximately 80 samples onwards. The findings support the conclusion that the PB-IAD approach outperforms the ML-based benchmark algorithm in dynamic manufacturing environments where data is limited. Furthermore, its overall performance remains consistently comparable to IF, despite it being trained on a substantial number of non-anomalous samples.

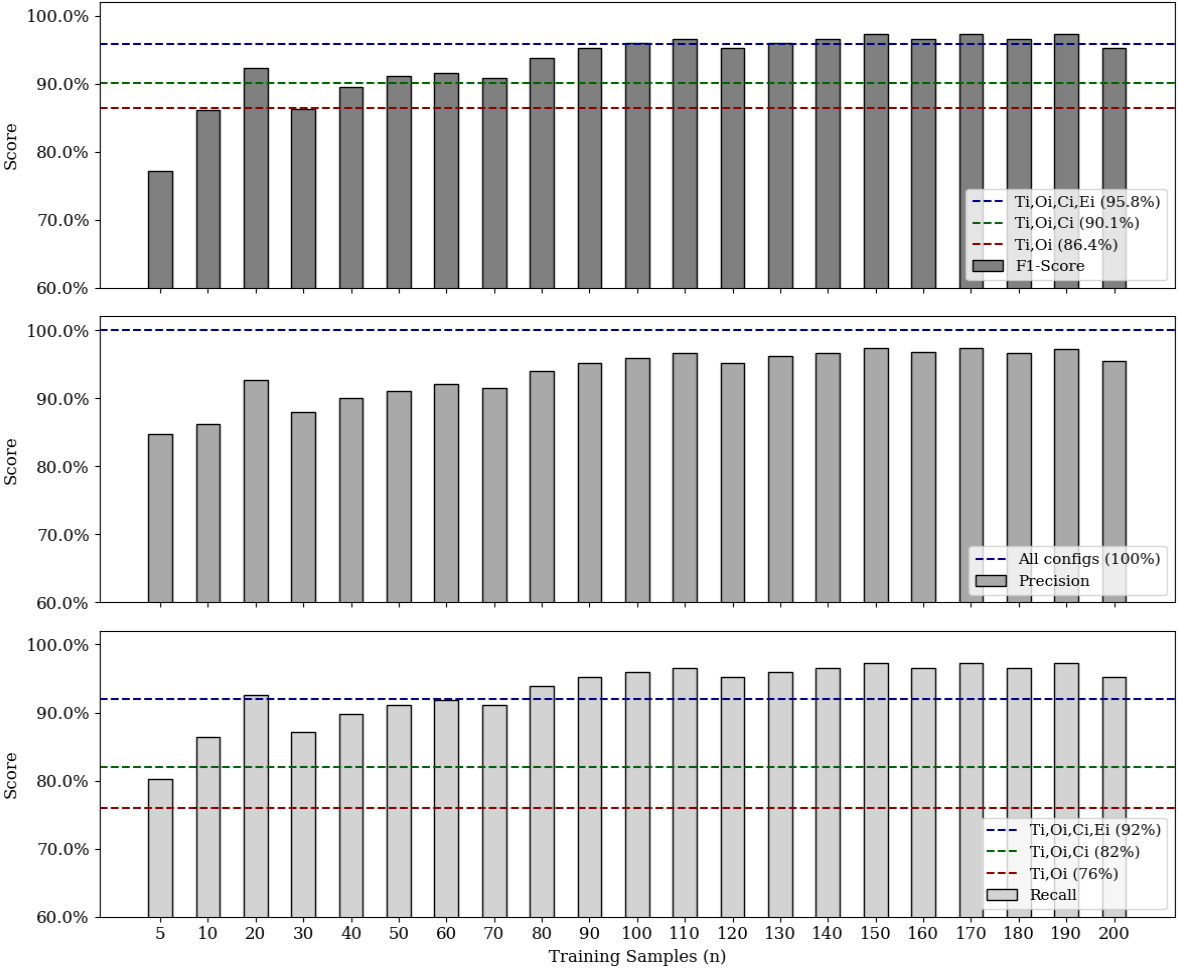

**Fig. 11.** Isolation Forest results of Scenario 3 benchmarked against PB-IAD.

## Conclusion and Future Work

Modern manufacturing environments are characterized by high dynamism, including frequent product changes, tool wear affecting processing conditions, and recurring ramp-up conditions. These factors often lead to process deviations and anomalies, which require rapid implementation of countermeasures to ensure product quality. In serial production settings, there is typically neither the time nor the expertise to collect extensive annotated datasets or to maintain complex data-driven approaches, particularly given that anomalies are, by nature, rare and unpredictable. As a result, frequent involvement of data scientists on the shopfloor becomes necessary, which limits the system's flexibility and responsiveness. To address these challenges, this paper introduces PB-IAD, a framework that leverages the perception capabilities of foundation models for industrial anomaly detection. The framework is designed with a prompt template and a pre-processing module that shifts responsibility from data scientists to process experts, allowing for flexible adaptability. By exploiting the extensive pretraining of foundation models, PB-IAD can operate based purely on semantic instructions, which is particularly valuable during the early ramp-up phases of production where annotated data is scarce or unavailable. When labelled data becomes available at a later stage, it can be incorporated as low-shot examples, further enhancing the framework's effectiveness. The experimental results demonstrate that PB-IAD, utilising GPT-4, attains competitive and, particularly in data-sparse scenarios, superior performance in anomaly detection tasks against algorithms like Isolation Forrest and PatchCore, relying solely on semantic instructions without the necessity for further training. These findings emphasise the advantage of utilizing foundation models with an in-context-learning approach for downstream tasks, such as industrial anomaly detection under real-world constraints. While this paper focuses on assessing the general capabilities of the proposed approach, future research should investigate the applicability of smaller, locally deployable models. Moreover, foundation models should not be viewed solely as standalone anomaly detectors, but as integral components of comprehensive quality management systems that directly interact with production machinery and support real-time troubleshooting. In this context, PB-IAD could be further extended within an agentic framework, analogous to the Vision-Language-Action (VLA) paradigm commonly employed in robotics applications. This would enable the system not only to perceive and reason about anomalies, but also to execute corrective actions directly on a machine.

## Declaration of generative AI and AI-assisted technologies in the writing process

During the preparation of this work the authors used generative AI and AI-assisted technologies in order to improve the language and readability of their paper. After using this tool/service, the authors reviewed and edited the content as needed and take full responsibility for the content of the publication.

## Acknowledgement

The research presented in this paper is part of the project "KIdoka: AI-based quality agent for autonomous quality control through proactive fault detection and troubleshooting in cable production" (DIK-2408-0033// DIK0633/01), which is funded by the Bavarian Ministry of Economic Affairs, Regional Development and Energy (StMWi) and supervised by "VDI/VDE-IT".

## Data availability

The MVTec dataset [51] from Scenario 1 is openly available. To fully support transparency and reproducible research, we have made the images of Scenario 2 (Stripped Wire) and the crimp force curves from Scenario 3 also publicly available through Zenodo [53,54].

# Supplementary Material

## A. Scenario specific context and expertise instruction prompts

> ## CONTEXT INSTRUCTION:
> The provided image shows a cross-sectional view of a multi-core electrical power cable.
> - The purpose is to **detect structural defects** caused by environmental or process-related factors.
> - The cable contains **three separate conductors**, each consisting of **stranded copper wires**.
> - Each conductor is **individually insulated** with coloured plastic:
>   - **Yellow-Green** (typically the protective earth conductor)
>   - **Blue** (typically the neutral conductor)
>   - **Brown or Grey** (typically the phase conductor)
> - The entire assembly is enclosed in a **white or grey outer insulating sheath** (main jacket).
> - The cable is viewed in cross-section, clearly showing the **conductor geometry, insulation, and copper strands**.
> - The images are collected from a **manufacturing environment**, where:
>   - **Lighting conditions**, **background**, **image scale**, and **zoom level** may vary.

**Fig. 12.** Context instruction of Scenario 1.

## EXPERTISE INSTRUCTION:
Non-Anomalous Condition:
- Each conductor (yellow-green, blue, brown/grey) is **present and visibly intact**.
- **Copper strands** are uniformly packed, untangled, and lie neatly within the insulation.
- **No strands are bent, protruding, or damaged**.
- **Insulation (inner and outer)** is smooth, unbroken, and free from cuts, cracks, or deformation.
- The **outer sheath** fully encases all conductors and shows no damage.
- **Conductors are centered and evenly spaced** within the sheath.
- Note: A centered and tidy appearance does **not guarantee normalcy** — strand integrity and insulation condition must be individually verified.

Anomalous Conditions (any of the following constitutes an anomaly):
- **Bent Wire**: One or more copper strands are visibly bent or protruding out of alignment.
- **Cable Swap**: Expected color-code (yellow-green, blue, brown/grey) is violated; incorrect or repeated colors appear.
- **Combined**: Multiple defect types (e.g., bent strands and insulation damage) occur in the same sample.
- **Cut Inner Insulation**: Visible damage (cuts, cracks, notches) to the colored plastic surrounding a conductor.
- **Cut Outer Insulation**: Visible breach or crack in the outer white/grey sheath.
- **Missing Cable**: One of the three expected conductors is absent.
- **Missing Wire**: One or more copper strands are missing inside the conductor.
- **Poke Insulation**: Copper strands are poking into or through the insulation layer.

Important Notes:
- Any **single anomaly** is enough to flag the image as faulty.
- **Lighting, focus, and angle may vary**, but defect visibility should be consistent based on structure and geometry.
- Automated detection must remain robust to variations in background and visual noise.

**Fig. 13.** Expertise instruction of Scenario 1.

## CONTEXT INSTRUCTION:
The provided images show individual **insulated electrical wires** that have been **partially stripped** to expose the **copper strand core**. These images are captured in a **manufacturing environment**, where variability in **lighting**, **background**, **zoom**, and **wire orientation** may occur.
- Each wire consists of **multiple fine copper strands** bundled together inside a **colored plastic insulation** (e.g., red, white, yellow, brown).
- The stripping process exposes the copper strands, which are expected to be clean, evenly aligned, and uniformly cut.
- Visual inspection aims to detect production defects caused during the **stripping or handling process**.

**Fig. 14.** Context instruction of Scenario 2.

## EXPERTISE INSTRUCTION:

Non-Anomalous Condition:

- The **copper strands** are fully intact, tightly bundled, and extend evenly from the insulation.
- No individual strand is missing, pulled out, or visibly shorter than the others.
- The insulation ends cleanly at the base of the bundle, without deformation or intrusion.
- The wire may be slightly curved or rotated, but **all strands must remain uniform and undamaged**.

Anomalous Conditions:

Any of the following visible issues is sufficient to classify the image as anomalous (1):

1. **Pulled Strand(s)**:
   - One or more copper strands are **pulled significantly longer** than the rest and protrude beyond the uniform strand length.
   - These strands appear detached from the main bundle alignment.
2. **Cut Strand(s)**:
   - One or more strands are **clearly shorter**, **partially cut**, or have **visible notches**.
   - Strands may end abruptly or be visibly damaged at the tips.
3. **Combined Anomaly**:
   - A combination of pulled and cut strands in the same image.

**Notes**:

- Minor variation in wire tilt or strand curvature is acceptable if the bundle remains consistent and undamaged.
- Do not flag minor surface discoloration or lighting-related visual artifacts unless accompanied by structural damage.

**Fig. 15.** Expertise instruction of Scenario 2.

## CONTEXT INSTRUCTION:

You are provided with **features** of **force-displacement curves**, which are recorded during a **wire crimping process** to monitor tool behavior and process quality.

- Each curve represents a **time series** where the x-axis is **displacement** and the y-axis is **force**.
- These measurements reflect the applied force during crimping and are used to detect potential process faults.
- **Three non-anomalous reference curves** are always provided alongside the **TEST SAMPLE**, allowing visual and structural comparison.
- The applied force may vary due to **tool wear**, **material inconsistency**, or **machine misbehavior**, all of which may cause deviations in curve shape, slope, or area.

**Fig. 16.** Context instruction of Scenario 3.

## EXPERTISE INSTRUCTION:

To distinguish between **non-anomalous (0)** and **anomalous (1)** samples, focus on **specific regions** of the curve rather than the entire sequence.

Non-Anomalous Curve Characteristics:
- The slope, shape, and force progression across the curve closely match the reference curves.
- In particular:
  - Between data points **150 and 190**, the force increases with a **similiar slope** as the reference curves.
  - Between data points **250 and 300**, the force maintains expected levels, resulting in a **comparable area under the curve (AUC)**.

Anomalous Curve Characteristics (any one is sufficient to flag anomaly):
1. **Steep Force SLOPE (150-190)**:
   - The test sample shows a **significantly steeper slope** in this region than any of the reference curves.
2. **Reduced Area under the Curve (250-300)**:
   - The test sample shows **lower force values**, producing a **smaller area under the curve** compared to non-anomalous references.

3. **Combined Effects**:
   - Both deviations may appear in the same curve, further confirming an anomaly.

**Notes**:
- Minor local noise is acceptable as long as overall behavior matches the reference trends.
- Classification is **structure-based**, not appearance-based—focus on curve geometry and statistical regions.

**Fig. 17.** Expertise instruction of Scenario 3.